\documentclass[fleqn,10pt]{wlscirep}
\usepackage[utf8]{inputenc}
\usepackage[T1]{fontenc}
\usepackage{lineno}
\usepackage{pgf}

\title{Oil Spill Drone: A Dataset of Drone-Captured, Segmented RGB Images for Oil Spill Detection in Port Environments}

\author[1,$\dag$]{Thomas De Kerf}
\author[1,$\dag$]{Seppe Sels}
\author[1,2]{Svetlana Samsonova}
\author[1]{Steve Vanlanduit}

\affil[1]{InViLab Research Group, Faculty of Applied Engineering, University of Antwerp}
\affil[2]{Port of Antwerp-Bruges}

\affil[*]{corresponding author(s): Thomas De Kerf (thomas.dekerf@uantwerpen.be)}

\affil[$\dag$]{these authors contributed equally to this work}

\begin{abstract}

The high incidence of oil spills in port areas poses a serious threat to the environment, prompting the need for efficient detection mechanisms. Utilizing automated drones for this purpose can significantly improve the speed and accuracy of oil spill detection. Such advancements not only expedite cleanup operations, reducing environmental harm but also enhance polluter accountability, potentially deterring future incidents. Currently, there's a scarcity of datasets employing RGB images for oil spill detection in maritime settings. This paper presents a unique, annotated dataset aimed at addressing this gap, leveraging a neural network for analysis on both desktop and edge computing platforms. The dataset, captured via drone, comprises 1268 images categorized into oil, water, and other, with a convolutional neural network trained using an Unet model architecture achieving an F1 score of 0.71 for oil detection. This underscores the dataset's practicality for real-world applications, offering crucial resources for environmental conservation in port environments.
\end{abstract}
\begin{document}

\flushbottom
\maketitle

\thispagestyle{empty}

\section*{Background \& Summary}
The topic of oil spills stands out in the discourse on environmental disasters, highlighted by extensive analyses in various seminal works~\cite{Beyer2016,Readman1996}. These oil spills have brought widespread devastation to ecosystems. While such large-scale spills often dominate media narratives, it is important to note that most oil spills are actually smaller in scale (under 700 tonnes) and tend to occur in or near ports. Recent data reveals that about 66\% of oil spills are of medium size, ranging from 7 to 700 tonnes, and more than half of these incidents are reported within the confines of port areas, as indicated by the latest oil tanker spill statistics~\cite{ITPOF2019}. This prevalence of medium-sized spills in such critical areas underscores the need for focused research and improved response strategies in port environments.

Presently, oil spill detection in ports relies largely on accidental discovery by port authority inspectors, leading to delays in reporting and clean-up. This delay is critical, as oil slicks in ports can move at speeds of 0.4 to 0.75 cm/s, potentially travelling over 2 km within an hour, making swift identification (within 30 minutes) crucial for the following benefits~\cite{Janati2020}. 
Due to the extended area that a port area can take up, using fixed cameras, is not a feasible option. Therefore drone based measurements are much more suitable as they can traverse a large area, autonomously and signal if there is an oil spill detected. There are implementations.
Additionally, the added detection rate has the following advantages:
\begin{itemize}[leftmargin=*,labelsep=5.8mm]
\item Reduced environmental harm due to quicker clean-up.
\item More efficient containment and clean-up of the oil.
\item Minimal disruption to port operations, reducing economic losses.
\item In some cases, clearer identification of the responsible polluter.
\end{itemize}

Past studies have predominantly concentrated on monitoring large-scale oil spills through satellite imagery and Synthetic Aperture Radar (SAR)~\cite{topouzelis2008oil,fiscella2000oil}techniques. While these methods are effective in open sea conditions, their effectiveness is compromised within port environments due to several limitations. SAR technology, in particular, is less efficient in ports where wave action is minimal and oil spills typically manifest as thin layers, not providing the significant dampening effect needed for SAR detection.

Furthermore, the use of satellite and high-altitude aerial imagery presents challenges in port settings. These methods, while beneficial for broad surveillance, do not offer the fine spatial resolution necessary for identifying smaller spills common in the intricate and bustling nature of port areas. Additionally, the financial implications of satellite usage are substantial, with the high costs becoming a barrier to the routine and detailed monitoring required for managing the complex dynamics of ports.

The integration of RGB and hyperspectral imaging technologies~\cite{salem2001hyperspectral,khanna2018comparing, rs15204950} has also been explored, yet these approaches introduce constraints in low-budget commercial scenarios. Given these considerations, there is a pressing demand for alternative strategies better tailored to the specific demands of port environments. These strategies must balance cost-efficiency with the capability to provide accurate, high-resolution monitoring for effective oil spill detection and management.

In our previous work~\cite{de2020oil}, we demonstrated the potential benefits of combining RGB and infrared imagery for oil spill detection. However, the practical application of this approach has been hindered by the high minimum requirements for thermal cameras, rendering it economically unviable at present. Addressing this challenge, this paper introduces a pioneering solution: the first annotated dataset of RGB images specifically captured in a port environment.

This dataset represents a significant advancement in the field of oil spill detection in a port environment. The dataset consists of 1268 images that are fully annotated into three categories: oil, water, and other. Offering an invaluable resource for training and validating detection models. While there exists a body of research and datasets~\cite{7812788, 8438999}, that focuses on the segmentation of ships and port infrastructure, these studies notably omit the consideration of oil spills within the port environment.

Utilizing this dataset, we have developed a segmentation model that showcases the practical viability of drone-based RGB imaging for oil spill detection. The model demonstrates an f1 score of 0.72, 0.91 and 0.75 for the oil, water and other category respectively. This marks an overall effective classification, underscoring the feasibility of this approach. Our findings reveal that drone-based RGB imaging, without the need for thermal imaging, is not only feasible but also efficient for oil spill detection in port environments.

This research opens up new avenues for cost-effective and accurate oil spill monitoring, leveraging the agility and extensive reach of drone technology combined with the simplicity and effectiveness of RGB imaging. The results from this study are poised to make a significant impact on environmental protection strategies in port areas, offering a ready-to-deploy solution for timely and reliable oil spill detection.


\section*{Methods}
\subsection*{Data Collection}
The dataset for this study was meticulously gathered using a drone (Dronematrix YACOB and DJI Mavic2) equipped with high-resolution cameras (4000*2250 and 3840*2160 pixels). The drone flights, conducted from 09/2021 to 09/2023, captured a wide array of images under different environmental conditions. This strategy guaranteed a varied and exhaustive dataset, mirroring a multitude of situations that might arise within a port context. The deliberate selection of the periods for capturing images, encompassing different months and times throughout the day, was meticulously planned. This ensured the inclusion of a wide spectrum of lighting scenarios, weather changes, and the array of operational tasks characteristic of a port setting.

To align with privacy and confidentiality guidelines, we implemented specific alterations to the raw data, a vital measure to facilitate the dataset's broad utilization without compromising privacy. Critical adjustments involved identifying and processing text~\cite{baek2019character}, people, and logos. All elements that could reveal individual identities and specific locations within the images. These identified areas were subsequently inpainted~\cite{suvorov2021resolution} to generate anonymized images, ensuring they remain suitable for image processing, without impacting its efficacy.

Furthermore, specific identifiable markings on ships and dockside were also blurred. This measure was taken to ensure that no sensitive information about the port's operations or the identities of the vessels within the port was disclosed. Such modifications are essential in maintaining the integrity and confidentiality of the operations within the port, as well as respecting the proprietary rights of shipping companies.

These modifications, while necessary for privacy and ethical considerations, were implemented in a manner that did not compromise the quality of the data for its intended use in oil spill detection. The alterations were carefully executed to ensure that the key elements necessary for identifying and analyzing oil spills – such as the texture, spread, and coloration of the oil against the water – remained intact and discernible in the images.

\subsection*{Data Annotation}
The annotation process for our dataset was meticulously carried out using the software platform CVAT (Computer Vision Annotation Tool), accessible at https://cvat.ai. 
To ensure comprehensive coverage and accuracy, the images were divided into several subsets. Each subset was then assigned to different members of the InViLab research group for initial annotation. This division of labour allowed for a more focused and detailed approach to the annotation process, as each researcher could concentrate on a manageable portion of the dataset, thereby enhancing the overall quality and consistency of the annotations.

After the initial annotation phase, a rigorous quality control process was implemented. Senior researchers within the group undertook a thorough review of the annotated images. This step was crucial to ensure the highest level of accuracy and reliability of the annotations. The senior researchers meticulously double-checked each annotation, making corrections and adjustments where necessary. This collaborative and multi-tiered approach to annotation and quality assurance ensured the quality of the dataset, with annotations that accurately represented the various aspects and characteristics of oil spills in a port environment.

Each pixel is categorised into one of either these categories: oil, water or other. The other class is a combination of ships, quays, buildings and air. The choice of such an encompassing super category is made because the goal of the dataset is to distinguish between oil and water, not to identify the different other elements in a port environment. An overview of three randomly selected images, together with the annotated regions can be found in Figure ~\ref{fig:annotated_images}:

\begin{figure}[ht]
\centering
\includegraphics[width=\linewidth]{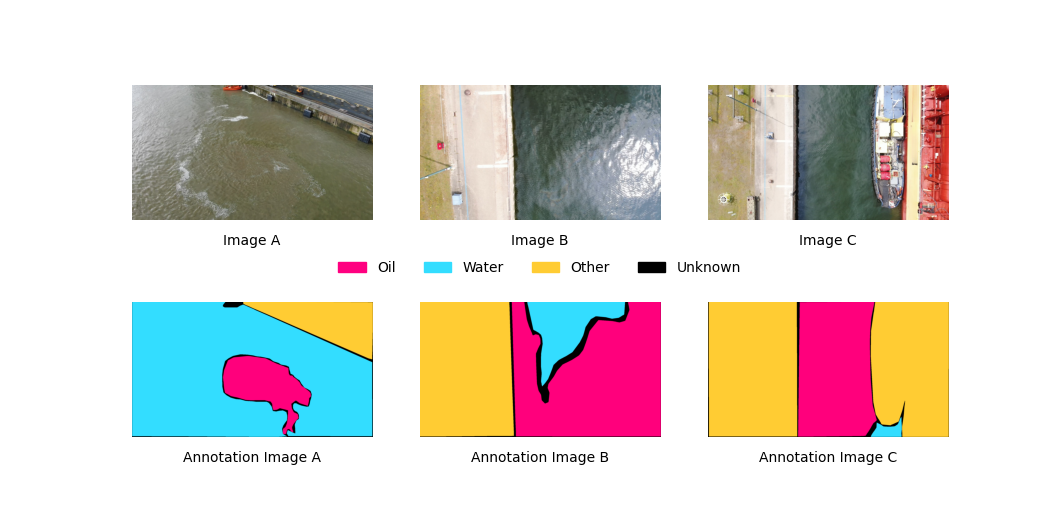}
\caption{Top row: Three randomly chosen images from the dataset. Bottom row: the annotated masks of the respective images.}
\label{fig:annotated_images}
\end{figure}

\subsection*{Data Records}
The dataset is publicly available and has been contributed to the open-source community, accessible via Zenodo~\cite{OilSpillDataset}. It comes pre-partitioned into training, testing, and validation subsets, following a 70/15/15 percentage split, respectively, to facilitate immediate use in machine learning workflows. Annotation data is provided in three formats to ensure compatibility with various training paradigms: the CamVid format~\cite{brostow2009semantic}, which was also employed during the network's training phase, as well as the widely recognized COCO~\cite{coco} and ImageNet~\cite{imagenet_cvpr09} formats. These multiple annotation formats enhance the dataset's versatility, allowing for seamless integration with different models and frameworks.

\subsection*{Dataset Statistics}
In Table \ref{tab:table_over}, we present essential statistics of our dataset, segmented into three key categories: Oil, Water, and other. This comprehensive overview is critical for understanding the scope and distribution of our data.

The first row of the table indicates the total number of images in which each category appears. The count of 994 images for Oil, 929 for Water, and 1106 for other reflects the dataset's diversity and the varying frequency of these elements within our collected imagery.

Moving to the second row, we detail the number of annotated pixels for each category. With 527,361,085 annotated pixels for Oil, 835,851,102 for Water, and 939,502,502 for other, these figures demonstrate the extensive level of detail that has been captured in our annotations. This granularity is pivotal for accurate analysis and application in related fields, such as machine learning and environmental monitoring.

The final row presents the percentage of relative annotations for each category, illustrating their proportional representation in the entire dataset. Here, Oil annotations constitute 22.9\%, Water 36.3\%, and Other 40.8\%. These percentages are indicative of the emphasis placed on each category within our dataset and underscore the balanced yet distinct focus on each element, providing a multifaceted perspective on the environmental subjects at hand.
\begin{table}[]
\centering
\caption{Dataset Overview by Category: This table provides a comprehensive breakdown of the dataset across three categories: Oil, Water, and Other. The first row indicates the total number of images featuring each respective category. The second row details the count of annotated pixels labelled under each category, reflecting the extent of detailed annotations provided. The final row presents the percentage of relative annotations per category, thereby illustrating their proportional representation in the entire dataset. }
\label{tab:table_over}
\begin{tabular}{@{}llll@{}}
\toprule
                                  & \multicolumn{3}{c}{\textbf{Category}}                   \\ \midrule
                                  & \textit{Oil} & \textit{Water} & \textit{Other} \\
\textit{\# Images}                & 994          & 929            & 1106                    \\
\textit{\# Annotated Pixels}      & 527361085    & 835851102      & 939502502               \\
\textit{Relative Annotation (\%)} & 22.9         & 36.3           & 40.8                    \\ \bottomrule
\end{tabular}
\end{table}

\subsection*{Neural Network Architecture}
We utilized a modified U-Net architecture~\cite{ronneberger2015unet} featuring an EfficientNet encoder for image segmentation tasks. This combination harnesses EfficientNet's efficient feature extraction capabilities and U-Net's precision in segmentation application~\cite{baheti2020eff}.

\textbf{Encoder:} The encoder is based on EfficientNet b4~\cite{tan2019efficientnet}, starting with a convolutional layer for initial feature extraction and followed by batch normalization. The core consists of Mobile Inverted Bottleneck Convolution Blocks with depthwise separable convolutions, incorporating Squeeze-and-Excitation layers for channel-wise feature recalibration and the MemoryEfficientSwish activation function for balance between performance and memory efficiency. The encoder was pretrained on the ImageNet Database~\cite{imagenet_cvpr09}.

\textbf{Decoder: }The U-Net decoder includes multiple DecoderBlocks with convolutional layers and ReLU activation, designed to upsample the feature maps to the original image size. Attention mechanisms are integrated to focus on salient features.

\textbf{Segmentation Head:} The final part of the network, the Segmentation Head, transforms decoded features into a segmentation map through a convolutional layer, followed by a sigmoid activation for binary classification.

\subsection*{Segmentation Results}
The table in \ref{tab:scores} offers a detailed evaluation of the multiclass image segmentation model used in our study. For each category - Oil, Water, and Other - the model's performance is assessed using four critical metrics: F1 Score, Precision, Recall, and Intersection over Union (IoU). These metrics collectively provide a multifaceted view of the model's accuracy and efficiency in correctly identifying and classifying each category within the dataset.

As observed, the model demonstrates strong performance across all categories, with particularly high scores in Precision and Recall for the 'Water' category. The F1 Score, which balances Precision and Recall, is notably high for 'Water' and 'Oil', indicating a robust capability of the model in handling these categories. The Intersection over Union (IoU) metric, which assesses the overlap between predicted and ground truth areas, also shows commendable results, especially for 'Water'.

\begin{table}[h]
\caption{This table provides a detailed breakdown of key performance indicators for the used multiclass image segmentation model. For each category, the table displays the F1 Score, Precision, Recall, and Intersection over Union (IoU), offering a comprehensive view of the model's effectiveness in accurately identifying and classifying each specific category within the images.}
\label{tab:scores}
\centering
\begin{tabular}{@{}lccc@{}}

\toprule
\multicolumn{1}{c}{\textbf{Metric}} & \multicolumn{3}{c}{\textbf{Category}}                   \\ \midrule
\textbf{}                           & \textit{Oil} & \textit{Water} & \textit{Other} \\
\textit{F1 Score}                   & 0.72         & 0.91           & 0.75                    \\
\textit{Precision}                  & 0.77         & 0.91           & 0.85                    \\
\textit{Recall}                     & 0.79         & 0.98           & 0.77                    \\
\textit{IoU}                        & 0.69         & 0.86           & 0.62                    \\ \bottomrule
\end{tabular}
\end{table}

\begin{figure}[h!]
\centering
\includegraphics[width = \linewidth]{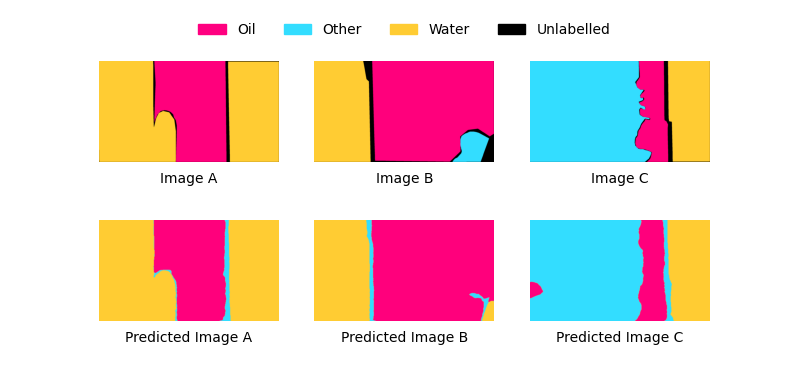}
\caption{Predictions on three images from the dataset. The top row represents the annotated mask, and the bottom row is the predicted mask by the neural network.}
\label{fig:predictions}
\end{figure}

While the model demonstrates accurate results according to the quantifiable metrics, a closer examination of the predictions, see Figure \ref{fig:predictions}, indicates certain limitations in its performance. In particular, the model appears to struggle with distinguishing between closely related categories, as evidenced by the misclassified regions in the predicted images. For instance, areas that should be classified as Water are occasionally marked as Oil, suggesting a potential confusion within the model when dealing with similar textures or color gradients.

Overall, while the model's general accuracy is commendable, these observed discrepancies underscore the need for further refinement. Enhancing the model could involve adjusting the model's architecture and hyperparameters to better capture the nuances between different categories.


\section*{Data Records}
The dataset is accessible at \url{https://doi.org/10.5281/zenodo.10555314}. It is systematically organized into six folders: three for the image sets (train, test, and validation) and three corresponding folders for annotations. All images and their respective annotation masks are saved in the .png format, ensuring consistency and ease of use.


\section*{Technical Validation}
Our dataset was carefully curated through the deployment of drones equipped with cameras, ensuring the capture of a wide array of images across varying environmental conditions and locations. To uphold privacy while retaining the dataset's efficacy for oil spill detection, specific alterations were implemented. This included the use of advanced inpainting techniques and selective blurring to anonymize identifiable features effectively, ensuring the resulting images remained highly relevant for analytical purposes.

\textbf{Annotation Validation:} The annotation process was a collaborative effort led by the InViLab research group, with equitable distribution of images among members for initial annotation. Dr. Sels played a pivotal role in reviewing and refining these annotations to guarantee precision. Post-annotation, the images were anonymized and further processed using inpainting to preserve the integrity of the dataset without compromising network performance. This careful anonymization process was meticulously reviewed by both Dr. Sels and Dr. De Kerf, ensuring the highest standards of data quality and reliability were maintained.

\textbf{Influence of the Inpainting Process}: To assess the impact of our anonymization technique on model performance, we conducted a comparative analysis. Initially, a model was trained using the original dataset, which had not been subjected to the anonymization process. This model's accuracy was validated against a distinct validation set to establish a performance benchmark. Subsequently, the same model was evaluated using the anonymized dataset, ensuring consistency in testing conditions. Remarkably, the comparative analysis revealed no discernible difference in the validation results between the anonymized and non-anonymized datasets. This finding demonstrates that our inpainting-based anonymization process does not compromise the model's ability to detect oil spills, affirming the integrity of our data processing methodology.



\section*{Code availability}
The data for this study was generated independently of any coding process. For reproducibility and further research, the complete codebase for our analysis, figures, and tables creation is available on GitHub at \url{https://github.com/thomasdekerf/OilSpillDataset_Analysis}. This repository also contains the code for training and evaluating our neural network, alongside the optimally trained model saved in a .pth file format, facilitating direct use and exploration by other researchers.
\bibliography{sample}

\section*{Acknowledgements}
The authors wish to thank the contribution of the Port of Antwerp-Bruges for the collaboration and images. This research is made possible thanks to the Belgian SPF Economy ETF:Phairywind Project. 

\section*{Author contributions statement}

Must include all authors, identified by initials, for example:
A.A. conceived the experiment(s), A.A. and B.A. conducted the experiment(s), C.A. and D.A. analysed the results. All authors reviewed the manuscript. 

\section*{Competing interests}
The authors declare that they have no competing interests or personal relationships that could have influenced
the work reported in this paper.

\end{document}